\documentclass[11pt]{article}
\usepackage[dvips]{graphicx}
\usepackage{textcomp}
\usepackage{amsbsy}
\usepackage{latexsym}
\usepackage[mathscr]{eucal}
\usepackage{amsfonts,amsmath,amsthm}
\usepackage{amssymb}
\usepackage[usenames]{xcolor}
\usepackage{graphicx}
\usepackage{caption}
\usepackage{subcaption}
\usepackage[normalem]{ulem}

\usepackage{float}

\usepackage{fullpage}

\def\rnew{\color{magenta}}
\def\cnew{\color{blue} }
\begin{document}
\bibliographystyle{plain}
\title
{Neural Network Approximation of Refinable Functions }
\author{ Ingrid Daubechies, Ronald DeVore,  Nadav Dym, Shira Faigenbaum-Golovin, Shahar Z.~Kovalsky, \\
Kung-Ching Lin,  Josiah Park, Guergana Petrova,  Barak Sober
\thanks{%
   This research was supported by the 
   NSF grants DMS 18-17603 (RD-GP), Tripods Grant CCF-1934904 (RD-KL-JP-GP), ONR Contract
  N00014-20-1-278 (RD-GP), THEORINET Simons award 814643 (ND).}
}
\hbadness=10000
\vbadness=10000

\newtheorem{lemma}{Lemma}[section]
\newtheorem{prop}[lemma]{Proposition}
\newtheorem{cor}[lemma]{Corollary}
\newtheorem{theorem}[lemma]{Theorem}
\newtheorem{remark}[lemma]{Remark}
\newtheorem{example}[lemma]{Example}
\newtheorem{definition}[lemma]{Definition}
\newtheorem{proper}[lemma]{Properties}
\newtheorem{assumption}[lemma]{Assumption}
%
\newenvironment{disarray}{\everymath{\displaystyle\everymath{}}\array}{\endarray}

\def\RR{\rm \hbox{I\kern-.2em\hbox{R}}}
\def\NN{\rm \hbox{I\kern-.2em\hbox{N}}}
\def\ZZ{\rm {{\rm Z}\kern-.28em{\rm Z}}}
\def\CC{\rm \hbox{C\kern -.5em {\raise .32ex \hbox{$\scriptscriptstyle
|$}}\kern
-.22em{\raise .6ex \hbox{$\scriptscriptstyle |$}}\kern .4em}}
\def\vp{\varphi}
\def\<{\langle}
\def\>{\rangle}
\def\t{\tilde}
\def\i{\infty}
\def\e{\varepsilon}
\def\sm{\setminus}
\def\nl{\newline}
\def\o{\overline}
\def\wt{\widetilde}
\def\wh{\widehat}
\def\cT{{\cal T}}
\def\cA{{\cal A}}
\def\cI{{\cal I}}
\def\cV{{\cal V}}
\def\cB{{\cal B}}
\def\cF{{\cal F}}
\def\cY{{\cal Y}}
\def\V{\text{{\bf Vec}}}

\def\Relu {{\rm ReLU}}

\def\cD{{\cal D}}
\def\cP{{\cal P}}
\def\cJ{{\cal J}}
\def\cM{{\cal M}}
\def\cO{{\cal O}}
\def\Chi{\raise .3ex
\hbox{\large $\chi$}} \def\vp{\varphi}
\def\lsima{\hbox{\kern -.6em\raisebox{-1ex}{$~\stackrel{\textstyle<}{\sim}~$}}\kern -.4em}
\def\lsim{\hbox{\kern -.2em\raisebox{-1ex}{$~\stackrel{\textstyle<}{\sim}~$}}\kern -.2em}
\def\[{\Bigl [}
\def\]{\Bigr ]}
\def\({\Bigl (}
\def\){\Bigr )}
\def\[{\Bigl [}
\def\]{\Bigr ]}
\def\({\Bigl (}
\def\){\Bigr )}
\def\L{\pounds}
\def\pr{{\rm Prob}}
\newcommand{\cs}[1]{{\color{magenta}{#1}}}
\def\ds{\displaystyle}
\def\ev#1{\vec{#1}}     
\newcommand{\lt}{\ell^{2}(\nabla)}
\def\Supp#1{{\rm supp\,}{#1}}
\def\R{\mathbb{R}}
\def\E{\mathbb{E}}
\def\nl{\newline}
\def\T{{\relax\ifmmode I\!\!\hspace{-1pt}T\else$I\!\!\hspace{-1pt}T$\fi}}
\def\N{\mathbb{N}}
\def\Z{\mathbb{Z}}
\def\N{\mathbb{N}}
\def\Zd{\Z^d}
\def\Q{\mathbb{Q}}
\def\C{\mathbb{C}}
\def\Rd{\R^d}
\def\gsim{\mathrel{\raisebox{-4pt}{$\stackrel{\textstyle>}{\sim}$}}}
\def\sime{\raisebox{0ex}{$~\stackrel{\textstyle\sim}{=}~$}}
\def\lsim{\raisebox{-1ex}{$~\stackrel{\textstyle<}{\sim}~$}}
\def\div{\mbox{ div }}
\def\M{M}  \def\NN{N}                  
\def\L{{\ell}}               
\def\Le{{\ell^1}}            
\def\Lz{{\ell^2}}
\def\Let{{\tilde\ell^1}}     
\def\Lzt{{\tilde\ell^2}}
\def\Ltw{\ell^\tau^w(\nabla)}
\def\t#1{\tilde{#1}}
\def\la{\lambda}
\def\La{\Lambda}
\def\ga{\gamma}
\def\BV{{\rm BV}}
\def\Ga{\eta}
\def\al{\alpha}
\def\cZ{{\cal Z}}
\def\cA{{\cal A}}
\def\cU{{\cal U}}
\def\argmin{\mathop{\rm argmin}}
\def\argmax{\mathop{\rm argmax}}
\def\prob{\mathop{\rm prob}}

\def\cO{{\cal O}}
\def\cA{{\cal A}}
\def\cC{{\cal C}}
\def\cS{{\cal F}}
\def\bu{{\bf u}}
\def\bz{{\bf z}}
\def\bZ{{\bf Z}}
\def\bI{{\bf I}}
\def\cE{{\cal E}}
\def\cD{{\cal D}}
\def\cG{{\cal G}}
\def\cI{{\cal I}}
\def\cJ{{\cal J}}
\def\cM{{\cal M}}
\def\cN{{\cal N}}
\def\cT{{\cal T}}
\def\cU{{\cal U}}
\def\cV{{\cal V}}
\def\cW{{\cal W}}
\def\cL{{\cal L}}
\def\cB{{\cal B}}
\def\cG{{\cal G}}
\def\cK{{\cal K}}
\def\cX{{\cal X}}
\def\cS{{\cal S}}
\def\cP{{\cal P}}
\def\cQ{{\cal Q}}
\def\cR{{\cal R}}
\def\cU{{\cal U}}
\def\bL{{\bf L}}
\def\bl{{\bf l}}
\def\bK{{\bf K}}
\def\bC{{\bf C}}
\def\X{X\in\{L,R\}}
\def\ph{{\varphi}}
\def\D{{\Delta}}
\def\H{{\cal H}}
\def\bM{{\bf M}}
\def\M{{\mathcal M}}
\def\bx{{\bf x}}
\def\bj{{\bf j}}
\def\bG{{\bf G}}
\def\bP{{\bf P}}
\def\bW{{\bf W}}
\def\bT{{\bf T}}
\def\bV{{\bf V}}
\def\bv{{\bf v}}
\def\bt{{\bf t}}
\def\bz{{\bf z}}
\def\bw{{\bf w}}
\def \span{{\rm span}}
\def \meas {{\rm meas}}
\def\rhom{{\rho^m}}
\def\diff{\hbox{\tiny $\Delta$}}
\def\EE{{\rm Exp}}
\def\lll{\langle}
\def\argmin{\mathop{\rm argmin}}
\def\codim{\mathop{\rm codim}}
\def\rank{\mathop{\rm rank}}

\def \MM{{\mathscr M}}

\def\argmax{\mathop{\rm argmax}}
\def\dJ{\nabla}
\newcommand{\ba}{{\bf a}}
\newcommand{\bb}{{\bf b}}
\newcommand{\bc}{{\bf c}}
\newcommand{\bd}{{\bf d}}
\newcommand{\bs}{{\bf s}}
\newcommand{\bff}{{\bf f}}
\newcommand{\bp}{{\bf p}}
\newcommand{\bg}{{\bf g}}
\newcommand{\by}{{\bf y}}
\newcommand{\br}{{\bf r}}
\newcommand{\be}{\begin{equation}}
\newcommand{\ee}{\end{equation}}
\newcommand{\bea}{$$ \begin{array}{lll}}
\newcommand{\eea}{\end{array} $$}
\def \Vol{\mathop{\rm  Vol}}
\def \mes{\mathop{\rm mes}}
\def \Prob{\mathop{\rm  Prob}}
\def \exp{\mathop{\rm    exp}}
\def \sign{\mathop{\rm   sign}}
\def \sp{\mathop{\rm   span}}
\def \rad{\mathop{\rm   rad}}
\def \vphi{{\varphi}}
\def \csp{\overline \mathop{\rm   span}}
%
%
\newcommand{\beqn}{\begin{equation}}
\newcommand{\eeqn}{\end{equation}}
\def\beginproof{\noindent{\bf Proof:}~ }
\def\endproof{\hfill\rule{1.5mm}{1.5mm}\\[2mm]}

\newenvironment{Proof}{\noindent{\bf Proof:}\quad}{\endproof}

\renewcommand{\theequation}{\thesection.\arabic{equation}}
\renewcommand{\thefigure}{\thesection.\arabic{figure}}

\makeatletter
\@addtoreset{equation}{section}
\makeatother

\newcommand\abs[1]{\left|#1\right|}
\newcommand\clos{\mathop{\rm clos}\nolimits}
\newcommand\trunc{\mathop{\rm trunc}\nolimits}
\renewcommand\d{d}
\newcommand\dd{d}
\newcommand\diag{\mathop{\rm diag}}
\newcommand\dist{\mathop{\rm dist}}
\newcommand\diam{\mathop{\rm diam}}
\newcommand\cond{\mathop{\rm cond}\nolimits}
\newcommand\eref[1]{{\rm (\ref{#1})}}
\newcommand{\iref}[1]{{\rm (\ref{#1})}}
\newcommand\Hnorm[1]{\norm{#1}_{H^s([0,1])}}
\def\int{\intop\limits}
\renewcommand\labelenumi{(\roman{enumi})}
\newcommand\lnorm[1]{\norm{#1}_{\ell^2(\Z)}}
\newcommand\Lnorm[1]{\norm{#1}_{L_2([0,1])}}
\newcommand\LR{{L_2(\R)}}
\newcommand\LRnorm[1]{\norm{#1}_\LR}
\newcommand\Matrix[2]{\hphantom{#1}_#2#1}
\newcommand\norm[1]{\left\|#1\right\|}
\newcommand\ogauss[1]{\left\lceil#1\right\rceil}
\newcommand{\QED}{\hfill
\raisebox{-2pt}{\rule{5.6pt}{8pt}\rule{4pt}{0pt}}%
  \smallskip\par}
\newcommand\Rscalar[1]{\scalar{#1}_\R}
\newcommand\scalar[1]{\left(#1\right)}
\newcommand\Scalar[1]{\scalar{#1}_{[0,1]}}
\newcommand\Span{\mathop{\rm span}}
\newcommand\supp{\mathop{\rm supp}}
\newcommand\ugauss[1]{\left\lfloor#1\right\rfloor}
\newcommand\with{\, : \,}
\newcommand\Null{{\bf 0}}
\newcommand\bA{{\bf A}}
\newcommand\bB{{\bf B}}
\newcommand\bR{{\bf R}}
\newcommand\bD{{\bf D}}
\newcommand\bE{{\bf E}}
\newcommand\bF{{\bf F}}
\newcommand\bH{{\bf H}}
\newcommand\bU{{\bf U}}
\newcommand\cH{{\cal H}}
\newcommand\sinc{{\rm sinc}}
\def\enorm#1{| \! | \! | #1 | \! | \! |}

\newcommand{\dm}{\frac{d-1}{d}}

\let\bm\bf
\newcommand{\bbeta}{{\mbox{\boldmath$\beta$}}}
\newcommand{\bal}{{\mbox{\boldmath$\alpha$}}}
\newcommand{\bbi}{{\bm i}}

\def\nnew{\color{Red}}
\def\mnew{\color{Blue}}
\def\wnew{\color{magenta}}

\newcommand{\dI}{\Delta}
\newcommand\aconv{\mathop{\rm absconv}}

\newcommand{\relu}{\mathrm{ReLU}}
\maketitle
\date{}
\begin{abstract}    
 In the desire to quantify the success of neural networks in deep learning and other applications, there is a great interest in understanding which functions are efficiently approximated by the outputs of neural networks. By now, there exists a variety of results which show that a wide range of functions can be approximated with sometimes surprising accuracy by these outputs.
For example, it is known that the set of functions  that  can be approximated   with exponential accuracy (in terms of the number of parameters used)  includes, on one hand, very smooth functions   such as polynomials and analytic functions (see e.g. \cite{E,S,Y}) and, on the other hand,  very rough functions such as the Weierstrass function (see e.g. \cite{EPGB,DDFHP}), which is nowhere differentiable.    
In this paper, we add to the latter  class of rough functions by showing  that  it  also includes refinable functions. Namely, we show that refinable functions  are approximated by the outputs of deep ReLU networks with a  fixed width and increasing depth with accuracy exponential in terms of their number of parameters.   
Our results apply to functions used in the standard construction of wavelets as well as to functions constructed via subdivision algorithms in Computer Aided Geometric Design.    
\end{abstract}

\section{Introduction}
\label{SS:refinable} 

Neural Network Approximation (NNA) is concerned with how efficiently a function, or a class of functions, is approximated by
the outputs of neural networks.   One overview of NNA is given in \cite{DHP} but there are other noteworthy expositions on this subject such as \cite{EPGB,Pinkus}.    
The main theme of
NNA is to understand for specific functions, or classes of functions, how fast the approximation error tends to zero as the number $n$ of parameters of the neural net grows.   In this paper, we prove bounds on the rate of NNA for univariate refinable functions (see \eref{refine1}) when using deep networks with ReLU activation.

We follow the notation and use the results in \cite{DHP} for neural networks.  In particular, we denote by $\Upsilon^{W,L}(\Relu;d,N)$   the set of outputs of a fully-connected neural network with width $W$, depth $L$, input dimension $d$, output dimension $N$, and the \textit{Rectified Linear Unit} (ReLU) as the activation function.    Since
we shall use deep networks for the approximation of univariate functions, we introduce the notation
\be
\label{sigman}
\Sigma_n:=
{\Sigma_n(C',C):=\Upsilon^{C',Cn}}(\Relu;1,1),\quad n\ge 1,
\ee
where  $C'$ and $C$ are fixed.   The set $\Sigma_n$ is a nonlinear parameterized set depending
on at most  $\tilde C n$  parameters, where $\tilde C=\tilde C(C',C)$ depends only on $C'$ and $C$.  The elements in $\Sigma_n$ are known to be Continuous Piecewise Linear (CPwL)    (vector valued) functions on $\R$.
While each $S\in \Sigma_n$ is determined by  ${\cal O}(n)$ parameters, the number of breakpoints of a given $S$ may be exponential in $n$. However, as shown in \cite{DSD}, not all CPwL functions
with an exponential number of breakpoints are in $\Sigma_n$, and indeed, the membership of  $S$   in $\Sigma_n$ imposes strong dependencies between
the linear pieces.

We consider a univariate function $\phi:\R\to\R$ which is refinable in the sense that there are constants $c_j\in\R$, $j=0,\dots,N$,  such that
\be 
\label{refine1}
\phi(x)=\sum_{j=0}^N c_j\phi(2x-j),\quad x\in\R. 
\ee
The sequence $\bar c:=(c_j)_{j=0}^N$ is called the {\it mask} of the refinement equation \eref{refine1}. Because of \eref{refine1}, refinable functions $\phi$ are self-similar.
Note that  the functions satisfying \eref{refine1} are not unique since, for example,   any multiple of $\phi$ also satisfies the same equation. However, under very minimal requirements
on the mask $\bar c$, there is a unique solution to \eref{refine1} up to scaling.

There is by now a vast literature on refinable functions (see for example \cite{CDM}) which derives  various properties of the function  $\phi$ from assumptions on the mask $\bar c$. In our presentation, we describe our assumptions as properties imposed on
$\phi$ and thus, if the reader wishes to know which properties of the mask will guarantee our assumptions, they must refer to the existing literature,  in paricular
\cite{CDM,DL,DL1}.

Refinable functions are of particular interest in approximation theory because they provide the natural framework for every practical wavelet basis in which the basic wavelets have bounded support. One and several dimensional refinable functions are also the underlying mathematical constructs in subdevision schemes used in Computer Aided Geometric Design (CAGD).

We rely heavily on the results and techniques from  \cite{DL,DL1}. To keep our presentation as transparent as possible, we only consider refinable functions that satisfy
the  two scale relationship \eref{refine1}.   There are various generalizations of \eref{refine1}, including the replacement of the dilation factor $2$ by $k$ as well as generalizations of the definition of refinablity to the multivariate settings where the dilation is given by general linear mappings (matrices).  Generalizations of the results of the present paper to these broader settings is left to future work.

We  next introduce the Banach space $C(\Omega)$ of continuous and bounded functions $f:\Omega\to\R$, defined on an interval 
$\Omega\subset \R$ (which can be all of $\R$), and the uniform norm
\be
\label{unorm}
\|f\|_{C(\Omega)}:= \sup_{x\in\Omega} |f(x)|,\quad f\in C(\Omega).
\ee
We consider  the linear operator $V=V_{\bar c}$ , $V:C(\R)\to C(\R)$, 
given by
\be
\label{defV}
Vg(x):=\sum_{j=0}^N c_jg(2x-j),  \quad g\in C(\R),
\ee
and its composition with itself $n$ times
\be
\label{defVn}
V^n g:=\underbrace{V\circ V\circ\ldots\circ V}_{n\ \text{{\rm times}}} g.
\ee

The main contribution of our article is described formally in the following theorem.
\begin{theorem}
	\label{T:maintheorem}
	Let $\bar c=(c_j)_{j=0}^N$ be any  refinement mask, let $g$ be any CPwL function 
	which vanishes outside of $[0,N]$, and let $V$ be the linear operator of \eqref{defV}.
	Then, the function $V^ng$ is in $ \Sigma_n=\Upsilon^{ C',Cn}(\Relu;1,1)$
	with $C,C'$ depending only on 
	$N$ 
	and the number $m$ of breakpoints of $g$.
	
	\bigskip
	
\end{theorem} 
As a corollary, under certain standard assumptions   on $\bar c$, we show in Section \ref{S:properties} that a refinable function $\phi$ can be approximated by the elements of $\Upsilon^{C',Cn}({\rm ReLU};1,1)$ with exponential accuracy. More precisely, 
we prove that under certain  assumptions  on the mask $\bar c$, the  normalized solution $\phi$ of \eref{refine1} satisfies the following for $n=1,2,\dots$

\be\label{error}
E_n(\phi):= \dist(\phi, \Upsilon^{C',Cn}(\Relu;1,1))_{C(\R)} \le \tilde C\lambda^n, 
\ee
where $0<\lambda<1$ and $\tilde C$ depend on the mask.

Our main vehicle for proving these results  is  the  {\it cascade algorithm} which 
is  used to compute $V^ng$.
We describe this algorithm in Section
\ref{s:preliminaries}.
Note that in \cite{DL} the term cascade algorithm was used more narrowly to indicate that as a consequence of \eqref{refine1}, the numerical values of $f(\ell2^{-j})$ could be computed easily from a few $f(k2^{-j+1})$, where $2k$ is close to $\ell$. 
Now  we are using this terminology  in a more general sense, including also what in \cite{DL} was given the more cumbersome name of \textit{two-scale difference equation}. Notice that the cascade algorithm cannot be directly implemented by ReLU NNs and so various modifications in this algorithm need to be made.
These are given in the proofs of the theorem as portrayed in Section \ref{S:main}.

We wish to stress here that some of the lemmas we use in our proofs may be applicable to other settings of NNA. 
In particular, we draw the reader's attention to the result of \S \ref{S:products} and its utilization, 
which show that in some cases we can describe a multiplication of two functions from $\Sigma_n$ as a function in $\Sigma_n$.

It is well-known that the solutions to refinement equations are used in the construction of wavelets with compact support (see \cite{D10}) and  in subdivision algorithms in 
CAGD.
In Section \ref{S:properties}, we discuss how our main  theorem can be combined with the existing theory of refinable functions and the existing convergence results for the cascade algorithm to prove that under standard conditions on $\bar c$, the solution $\phi$ to the refinement equation can be approximated to exponential accuracy by the elements of $\Sigma_n$.
Finally,  in Section \ref{S:NNA}, we discuss how our results relate to $n$-term wavelet approximation.

\section{Preliminaries}\label{s:preliminaries}

In this section, we touch upon  some of the necessary tools that  describe the cascade  algorithm as outlined in the works of Daubechies-Lagarias \cite{DL,DL1}.

\subsection  {The operator $V$} 

We consider the action of $V$ on any continuous function $g$ supported on $[0,N]$.   It is hard to do a direct  analysis of $Vg$  because the points $2x-j$, $j=0,\dots,N$, are spread out.
However, note two important facts.
The first is that  the points appearing
in \eref{defV} are all equal to $2x$ modulo one. This means that there is at most one point from each interval $[j,j+1]$ and all these points differ by an integer amount. Secondly, 
since  $g$ is supported on $[0,N]$, only the points $2x-j$ that land in $[0,N]$ appear in \eref{defV}.  More precisely, the following  statement about $Vg$ holds. 
\begin{remark}
	\label{R:Vcpwl}
	If    $g:\R\to\R$ is a CPwL  function supported on $[0,N]$, then 
	$Vg$ is also a CPwL  supported on $[0,N]$.  Moreover, each breakpoint
	$\xi '$  of $Vg$ satisfies  $2\xi '= j+\xi$, $j=0,\ldots,N$, where $\xi$ is a breakpoint of $g$. In particular, given a CPwL function $g_0$ supported on $[0,N]$ with  breakpoints at the integers $\{0,\dots, N\}$, $V^n g_0$ has breakpoints at $j/2^n$,  $j = 0,1,\dots, N 2^n$.
\end{remark}

\noindent
Indeed, the fact that $Vg$ is supported on $[0,N]$ follows from the observation that each of the functions 
$g(2x-j)$, $j=0,\dots,N$, is
supported on $[j/2,(N+j)/2]\subset[0,N]$.

\bigskip   
It follows  from Remark \ref{R:Vcpwl}
that  if $g_0$ is a CPwL function supported on $[0,N]$ and  has breakpoints at the integers $\{0,\dots, N\}$, then $V^ng_0$ is a CPwL supported on $[0,N]$ with breakpoints $j/2^n$,  $j = 0,1,\dots, N 2^n$, and as
such,  $V^ng_0\in\Upsilon^{3,N2^n}(\Relu;1,1) $.
We shall  show that $V^ng_0$ is actually an output of an NN with much smaller depth.

In going forward, we put ourselves in the setting of the works of Daubechies-Lagarias \cite{DL,DL1}, where 
a better understanding of $Vg$ is facilitated by the introduction of the operator $\V$. It assigns to each $g\in C(\R)$ a 
vector valued function 
$G:=\V(g)$,  where $G:=(g_1,\dots,g_N)^T$ with
\be
\label{vectorg}
g_k(x):=g(x+k-1), \quad   x\in \R,\ k= 1,2,\ldots,N.
\ee
Even though   $\V(g)$ is defined on all of $\R$, we are mainly concerned with its values on $[0,1]$.   Note that the solution $\phi$ of \eqref{refine1} turns out to be supported on $[0,N]$. Therefore,   knowing the restriction to $[0,1]$ of $\V(\phi)$ is equivalent to knowing $\phi$ on its full support.
On that interval, 
$g_k$ is the piece of $g$ living on $[k-1,k]$, reparameterized to live on $[0,1]$. Note that 
\be
\label{contG}
g_{k}(1)=g_{k+1}(0),\quad k=1,\ldots,N-1.
\ee
Further, we 
define for $x\in \R$ and $n\geq 1$
\be
\label{vectorV}
G_n(x):= \V(V^ng)(x):= (V^ng(x),\dots,V^ng(x+N-1))^T,
\ee 
Before describing the cascade  algorithm which represents $G_n$ via bit extraction,   
we recall in the next subsection how we find the binary bits of a number $x\in [0,1]$.

\subsection{Binary bits and quantization}
\label{SS:binary}

Any $x\in [0,1]$,  can be represented  as
\be 
\nonumber
x=\sum_{k=1}^\infty B_k(x)2^{-k},
\ee
where the bits  $B_k(x)\in \{0,1\}$.  While such a representation  of $x$ is not unique, we shall use one particular representation where the bits are  found using the quantizer function
\be 
\nonumber
Q(x):=\chi_{[1/2,1]}(x),\quad x\in [0,1],
\ee 
with  $\chi_I$ denoting the characteristic function of a set $I$.
The first bit  of $x$ and its residual
are defined as  
\be
\begin{split}
	B_1(x)&=Q(x),  \\
	R(x):&=R_1(x):=2x-B_1(x) \\
	&=2x-Q(x)\in [0,1], 
\end{split}
\label{residual}
\ee
respectively.
The graph of $R$  has two linear pieces, one on $[0,1/2)$ and the other on $ [1/2,1]$ and a jump
discontinuity at $x=1/2$.   Each linear piece for $R$ has slope $2$.

\begin{figure}[]
	\centering
	
	\begin{subfigure}[b]{0.4\textwidth}
		\includegraphics[width=\textwidth]{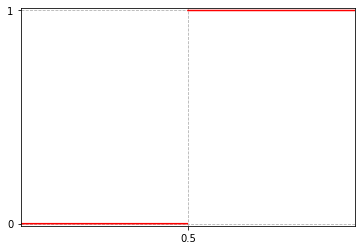}
		\caption{The graph of $Q$.}
	\end{subfigure}
	\hspace{5 mm}
	\begin{subfigure}[b]{0.4\textwidth}
		\includegraphics[width=\textwidth]{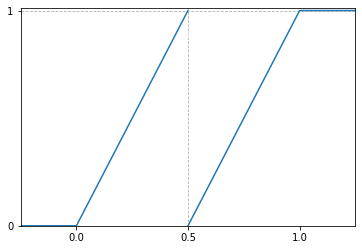}
		\caption{The graph  of $R$.}
	\end{subfigure}

	\caption{The graphs of $Q$ and $R$.}
	\label{fig:QR_plot}
\end{figure}

While for the most part we consider $R(x)$ only for $x\in[0,1]$, there are occasions where we need $R$ to be defined for  $x$ outside this interval.
For such $x$, we define $R(x):=0$ when $x\le 0$ and 
$R(x):=1$ when $x\ge 1$.   Figure \ref{fig:QR_plot} shows the graphs of $Q$ and $R$. 


\begin{figure}
	\centering
	
	\includegraphics[width=0.5\textwidth]{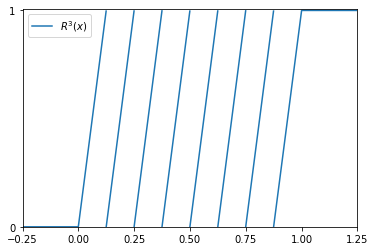}
	\caption{Graph of $R^3$.}
	
	\label{fig:Rj_plot}
\end{figure}

We find the later bits and residuals  recursively
as 
\begin{equation}
	\label{bits}
	\begin{split}
		B_j(x)&=Q(R_{j-1}(x)),\\
		R_j(x):&= 2R_{j-1}(x)-B_j(x)\\
		&=
		2R_{j-1}(x)-Q(R_{j-1}(x)),
		\quad j=2,3, \ldots.
	\end{split}
\end{equation}
Note that on $[0,1]$
\be
\label{new}
R^j:=\underbrace{R\circ \ldots\circ R}_{j\text{\,\,{\rm times}}}=R_j, \quad 
B_{j+1}=B_1\circ R^j, \quad j=1,2,\ldots .
\ee
The $R_j$'s are piecewise linear functions with jump discontinuities at the dyadic integers $k2^{-j}$, 
$k=1,2,\ldots,2^j-1$, see for example, Figure \ref{fig:Rj_plot} for the graph of $R^3=R_3$.
Note that, as in the case of $R_1$, we define 
$R_j(x):=0$ for $x\leq 0$ and 
$R_j(x):=1$ for $x\geq 1$. Then we will have that 
$R^j=R_j$ on the whole real line.

\subsection{The cascade algorithm}   
\label{SS:cascade}
We now  look at the computation of  $G_1= ((Vg)_1,\dots,(Vg)_N)^T$ on $[0,1]$ for general continuous functions $g$ supported on $[0,N]$.   Since
for $ x\in [0,1]$, and $ k=1,\ldots,N $ we have%
\be
\label{VG}
\begin{split}
	(Vg)_k(x):&=(Vg)(x+k-1)\\
	&=\sum_{j=0}^Nc_j g(2x+2k-2-j)  
	,\end{split}
\ee
we can  write 
\be
\begin{split}
	\label{VGwrite}
	g(2x+2k-j-2)&=g(R(x)+Q(x)+2k-j-2)\\
	&=
	\begin{cases}
		g_{2k-j-1}(R(x)),\quad  x\in[0,1/2),\\ \\
		g_{2k-j}(R(x)),\quad\quad x\in[1/2,1].
	\end{cases}
\end{split}
\ee
In this way, we get two different formulas depending on whether   $x\in [0,1/2)$ or $x\in [1/2,1]$.  For example, when $x\in [0,1/2)$,  using  the fact that $c_k=0$ if $k$ is not in $\{0,\dots,N\}$ and that $g_j(x):=g(x+j-1)=0$ for $x\in[0,1]$ when $j\leq 0$ or
$j\geq N+1$, we have for $ k=1,\ldots,N$,

\begin{equation*}
	\begin{split}
		(Vg)_k(x)&= \sum_{j=0} ^Nc_{j}g_{2k-j-1}(R(x))\\
		&= \sum_{j=2k-N-1}^{2k-1} c_{2k-j-1} g_j(R(x))\\
		&= (T_0G(R(x)))_k, \quad x\in [0, 1/2),
	\end{split},
\end{equation*}

where $T_0$ is the $N\times N$ matrix with $(i,j)$-th entry equal to $c_{2i-j-1}$,
\be
\label{T0}
T_0=(c_{2i-j-1})_{ij}, \quad i,j=1,\ldots,N.
\ee
A similar derivation gives 
$$
(Vg)_k(x)=(T_1G(R(x)))_k, \quad x\in [1/2,1],\quad k=1,\ldots,N,
$$
where now $T_1$ is the $N\times N$ matrix with $(i,j)$-th entry equal to $c_{2i-j}$,
\be
\label{eq:T1}
T_1=(c_{2i-j})_{ij},\quad i,j=1,\ldots,N.
\ee
More succinctly, we have for $ x\in[0,1] $
\be
\label{cascade1}
G_1(x)=\V(Vg) (x)= T_{Q(x)}G(R(x))=
T_{B_1(x)}G(R(x)).
\ee
Then, using \eqref{bits} we get
\begin{eqnarray}
	\nonumber
	G_2(x)&=&\V(V(Vg))(x)\\
	&=& T_{B_1(x)}\V(Vg)(R(x))\nonumber \\
	&=&
	T_{B_1(x)}T_{B_1(R(x))}G(R^2(x))\\
	&=&
	T_{B_1(x)}T_{B_2(x)}G(R^2(x)),\quad x\in[0,1],
	\nonumber
\end{eqnarray}
and if  we iterate this computation we get the cascade algorithm. Since $B_1(R^j(x))=B_{j+1}(x)$, we have for $x\in[0,1], \ n=1,2,\dots$
\be
\label{cascade}
G_{n}(x)= T_{B_1(x)}\cdots T_{B_n(x)}G(R^n(x)) .
\ee
It is useful to keep in mind what $T_{B_n(x)}$ looks like as $x$ traverses $[0,1]$. It alternately takes the values $T_0$ and $T_1$
with the switch coming at the dyadic integers $j2^{-n}$,  $1\le j<2^n$.
%



\section{Proof of the main theorem}
\label{S:main}  

Before we start proving Theorem \ref{T:maintheorem}, we observe a simple fact regarding how $V^n$ behaves with respect to the translation operator.
This fact, which is described in the following lemma, will help us simplify the proof of Theorem \ref{T:maintheorem}.


\begin{lemma}
	\label{L:shift}  Let $g$ be a continuous function on $\R$  
	and for any $\delta$, consider the translated function
	%
	$$
	\tilde g(\cdot ):=g(\cdot-\delta).  
	$$
	Then,  for each $n\ge 1$,
	\be
	\label{Vtranslate}
	V^n( g)(x)= [V^n(\tilde g)] (x+2^{-n}\delta),\quad x\in \R.
	\ee

\end{lemma}
\noindent
{\bf Proof:}  Let us first see the action of $V$ on $g$.   Since $g(x)= \tilde g(x+\delta)$, we have
\begin{eqnarray}
	\label{Vt1}
	Vg(x)&=&\sum_{k=0}^Nc_kg(2x-k)\nonumber\\
	&=&\sum_{k=0}^Nc_k \tilde g(2x-k+\delta)\\
	&=& \sum_{k=0}^Nc_k \tilde g(2(x+\delta/2)-k)\nonumber\\
	&=&[V\tilde g](x+\delta/2)\nonumber.
\end{eqnarray}
This proves the case $n=1$ in \eref{Vtranslate}.  
We next complete the proof  of \eref{Vtranslate} for all $n\ge 1$ by induction.  Suppose that we have established the result for a given value of $n$.  Consider the function
$h:= V^n(g)$.  Formula \eref{Vtranslate} says that 
$\overline h:= V^n(\tilde g)$  satisfies
\be
\label{satisfies}
\overline h(x)=h(x- 2^{-n}\delta ), \quad x\in \R. 
\ee
So we can apply \eref{Vt1} with $h$ in place of
$g$ and   obtain
\begin{eqnarray}
	V^{n+1}g(x)&=&V(h)(x)\nonumber\\
	&=& V(\overline h) (x+2^{-n-1}\delta )\nonumber\\
	&=& V^{n+1}(\tilde g) (x+2^{-n-1}\delta)\nonumber.
\end{eqnarray}
This advances the induction and proves   \eref{Vtranslate}.\hfill $\Box$

We turn now to discuss the proof of Theorem \ref{T:maintheorem}.  We first show how to prove the theorem when  $g=g_0$, where $g_0$ is  any  CPwL function that has support in $[0,N]$ and has breakpoints
at the integers.  We make this assumption on the breakpoints $g$ only for transparency of the proof.  We remark later how the same method of proof gives the theorem for arbitrary CPwL functions $g$.

We represent  $g_0$ as a linear combination of hat functions each of which has a translate with  support in $[1/8,7/8]$. 
The fact that these functions are supported on this sub-interval will be pivotal in many of the lemmas and theorems below and we will therefore use the following terminology throughout the paper.
We call a univariate function $g$ \textit{special} if:
\begin{itemize}
	\item $g$ is a non-negative CPwL function defined on $\R$.
	\item  the support of $g \subset  [1/8,7/8].$
\end{itemize}
Therefore, the translates of the hat functions in the representation of $g$ are special functions and all our results below, proven for special functions, can be applied to these translates.   

\begin{figure}
	\centering
	
	\includegraphics[width=0.5\textwidth]{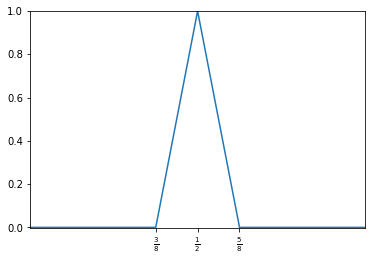}

	\caption{The graph of ${H}:=H_4$.}
	\label{fig:H_plot}
\end{figure}

Let $H$ be the 'hat function' with break points at $\{3/8,1/2,5/8\}$ and 
$H(\frac{1}{2})=1$, see Figure \ref{fig:H_plot}. That is, $H$ is given by the formula
\begin{eqnarray}
	\label{gg}
	H(x)=\begin{cases}
		0, \quad \quad \quad \quad x\leq \frac{3}{8}, \,\,x\geq \frac{5}{8},\\
		8x-3,\quad \quad \frac{3}{8}\leq x\leq \frac{1}{2},\\
		-8x+5,\quad \frac{1}{2}\leq x\leq \frac{5}{8}.
	\end{cases}
\end{eqnarray}
Clearly $H$ is a special function with $m=3$ break points. 
Although the function under investigation $g_0$ is not special, it can be written as a linear combination of shifts of the special function $H$,
\be
\label{repf}
g_0(x)=\sum_{j=1}^{8N-1}
g_0(j/8)H_j(x), \quad  
H_j(\cdot):= H(\cdot-\frac{j-4}{8}).
\ee  
Therefore, from \eref{repf}, Lemma \ref{L:shift} and the fact that $V^n$ is a linear operator it follows that
\begin{eqnarray}
	\label{newform}
	V^n g_0(x)&=&\sum_{j=1}^{8N-1}
	g_0(j/8)V^n(H(\cdot -\frac{j-4}{8}))(x)\nonumber\\
	&=&
	\sum_{j=1}^{8N-1}
	g_0(j/8)V^nH(x-2^{-n-3}(j-4)).  
\end{eqnarray}
Accordingly, the discussion in the following subsections concentrate on special functions.
We will come back to finalize the proof of the main theorem in the closing subsection.

\subsection{The function $g(R^n)$ is an output of an NN for special functions $g$}
In this section, we shall show that for certain  choices of $g$, the function  $g(R^n)$ is in the set $\Upsilon^{C,n+1}(\Relu;1,1)$,
where $C$ depends only on the mask $\bar c$ and the function $g$. 
If  $g$ is a 
special function, then the 
corresponding vector 
function $G$, viewed as a function on $[0,1]$, is 
$$
G=\V(g)=( g,0,\ldots,0)^T.
$$
Namely, all its coordinates are zero except the first one, which is the nonzero function $g$ supported 
on $[1/8,7/8]$.
Therefore $G(R^n(x))=(g(R^n(x)),0,\ldots,0)^T$, when considered as a function on $[0,1]$,  and is the zero vector when $R^n(x)$ takes a value  outside $[1/8,7/8]$.  Since the formula for $R^n$ is $R^n(x)=2^nx-k$ on $[k2^{-n},(k+1)2^{-n}]$, $0\le k<2^n$, this gives
\begin{equation}\label{know10}
	g(R^n(x)) = 0, \quad x\in  \Lambda,
\end{equation}
where
\begin{equation*}
	\Lambda := [0,1]\cap \bigcup_{0\le j\le 2^n} [j2^{-n}-2^{-n-3},j2^{-n}+2^{-n-3}].
\end{equation*}
In particular, the support of $g(R^n(x))$ is contained in 
$[2^{-n-3},1-2^{-n-3}]$.

The first step in our argument to prove that  $g(R^n)$ is an output of an NN is  to replace the discontinuous functions $R^j$ by  the  CPwL function $\hat R^j:=\hat R^j_{\alpha,\beta}$.  We shall give a family of possible replacements which depend on the choice
of two parameters $\alpha,\beta$ satisfying
\be
\label{choice} 7/16<\alpha <\beta< 1/2.
\ee

\vskip .1in

\begin{figure}
	\centering
	\begin{subfigure}[b]{0.475\textwidth}
		\centering
		\includegraphics[width=\textwidth]{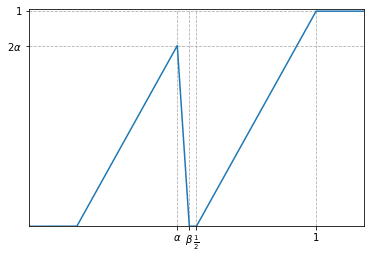}
		\caption{Graph of $\hat R_{\alpha,\beta}$.}
		
	\end{subfigure}
	\hfill
	\begin{subfigure}[b]{0.475\textwidth}
		\centering
		\includegraphics[width=\textwidth]{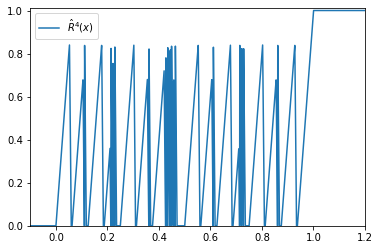}
		\caption{Graph of $\hat R^4_{\alpha,\beta}$.}
		
	\end{subfigure}
	
	\caption{The graphs of $\hat R_{\alpha,\beta}$ and its composition $\hat R^4_{\alpha,\beta}$ for $\alpha=0.42$, $\beta = 0.47$. 
	}
	\label{fig:R_ab_plot}
\end{figure}

\noindent
{\bf Definition of $\hat R_{\alpha,\beta}$:}  {\it   We let $\hat R:=\hat R_{\alpha,\beta}$ be the CPwL function defined on $\R$, see Figure {\rm \ref {fig:R_ab_plot}}(a),
	with breakpoints at $\{0, \alpha, \beta, 1/2,1\}$ which satisfies:
	\begin{itemize}
		\item $\hat R(x)=0$,    $x\le 0$;
		\item $\hat R(x)=R(x)$,  for $0\le x\le \alpha$ and for $x\ge 1/2$;
		\item  on $[\alpha,\beta]$, $\hat R$ is the linear function that interpolates $R(\alpha)$ at $\alpha$ and interpolates $0$ at $\beta$;
		\item  $\hat R(x)=0$ on $[\beta,1/2]$.
	\end{itemize}
}

In the next lemma, we summarize the properties of  $\hat R^n:=\underbrace{\hat R\circ \ldots\circ \hat R}_{n\text{\,\,{\rm times}}}$
that we will need in going forward.  For its statement, we introduce for $\delta:=1/2-\alpha$  the sets
\begin{eqnarray}
	\label{defEj} E_j&:=&\bigcup_{0<i <2^j}[i2^{-j}-\delta2^{-j+1}, i2^{-j}], \quad j=1,2,,\dots,\nonumber\\
	\Omega_k&:=&[0,1]\setminus \bigcup_{j=1}^kE_j,\quad k=1,2,\dots .
\end{eqnarray}

\begin{lemma}
	\label{L:Rhat} For each $n\ge 1$, the  function $\hat R^n$ has the following properties:
	\vskip .1in
	\noindent
	{\rm (i)} $\hat R^n\in \Upsilon^{5,n}(\Relu;1,1)$;
	\vskip .1in
	\noindent
	{\rm (ii)} $\hat R^n(x)=R^n(x)$ for  $x\in \Omega_n$;
	\vskip .1in
	\noindent
	{\rm (iii)} For $\delta':=1/2-\beta$, we  have $\hat R^n(x)=0$, at least at every $x\in [k2^{-n} -\delta' 2^{-n+1},k2^{-n}]$, $0<k<2^n$.
\end{lemma}
\noindent
{\bf Proof:}  Since $\hat R$ is in $\Upsilon^{5,1}(\Relu;1,1)$ we have that the composition $\hat R^n$ is in $\Upsilon^{5,n}(\Relu;1,1)$, see page 10  in \cite{DHP}.  This proves (i).  

We   prove (ii) by induction on $n$.  From the definition of $\hat R$, we have that $\hat R=R$ except for the interval $E_1=[1/2-\delta,1/2]$ and therefore we have
proved the case $n=1$.   Suppose we have established (ii) for a value of $n\ge 1$.   Let $x\in  \Omega_{n+1}$.  Then, $\hat R^n(x)= R^n(x)$ because $\Omega_{n+1} \subset \Omega_{n}$.     Therefore,  we only need to check that $\hat R (y)=R(y)$  when  $y=R^n(x)$, $x\in \Omega_{n+1}$.   Any such $x$ is in
an interval $[k2^{-n},(k+1)2^{-n}]$ for some  $0\le k<2^n$ and $y=R^n(x)=2^nx-k$.  If $R(y)\neq \hat R(y)$ then $y\in [1/2-\delta,1/2]$ and therefore $x\in E_{n+1}$.   Therefore, we have proved (ii). 

Finally, we prove (iii) also by induction on $n$. This statement is clear from the definition of $\hat R$ when $n=1$.  Suppose that we have proved (iii)
for  a value of $n\ge 1$ and consider the statement for $n+1$.   We consider two cases depending on the parity of $k$.  If $k=2j$ is even, then 
the interval under consideration is $2^{-n-1} [2j-2\delta', 2j] \subset 2^{-n}[j-2\delta',j]$.   Hence, we know $\hat R^n(x)$ vanishes on   this interval and since $\hat R(0)=0$, we get that $\hat R^{n+1}$ also vanishes on this interval.   Consider now the case that $k=2j+1$. Then,  the interval under consideration is 
\begin{eqnarray}
	2^{-n-1}[2j+1-2\delta',2j+1]&=&2^{-n} [j+\beta, j+1/2]\textbf{}\nonumber\\
	&\subset& 2^{-n} [j, j+2\alpha]\nonumber\\
	&\subset& \Omega_n,\nonumber
\end{eqnarray}
since $\alpha>7/16>1/4$. For $x$ in that interval, because of (ii), we have  
$$
\hat R^n(x)=R^n(x)= 2^nx-j\in [1/2-\delta',1/2].
$$
Hence,
$\hat R^{n+1}(x)=\hat R(\hat R^n(x))=0$.
\hfill $\Box$

\begin{remark}
	Note that in addition to the assumption $7/16<\alpha<\beta<1/2$, we require that $1/2-2^{-n}<\alpha$ so that the sets  $E_j$, $j=1,\ldots,n$, do not overlap. 
\end{remark}

\begin{remark}
	\label{R:fill}
	In the construction of NNs in this paper, we usually  construct NNs whose layers have different width. However, we can always add additional nodes to these layers so that we end up with a fully-connected NN with all layers being the same size.  
\end{remark}

Now, we are ready to state and prove the main theorem in this section.

\begin{theorem}
	\label{T1}
	Let $g$ be a special function that  has  at most $m\ge 3$ breakpoints.  Then, for any $n\ge 1$,
	$$
	g(R^n(x))\in \Upsilon^{2\max\{m,5\},n+1}(\Relu;1,1).
	$$
\end{theorem}
\noindent
{\bf Proof:}  We choose $7/16<\alpha_1<\beta_1<\alpha_2<\beta_2<1/2$.           Let us denote by $\hat R_1$ the function $\hat R_{\alpha_1,\beta_1}$ and denote by $\hat R_2:=\hat R_{\alpha_2,\beta_2}$.  We claim that
\be
\label{claim11}
g(R^n(x))=\min\{ g(\hat R_1^n(x)),g(\hat R^n_2(x))\},\quad x\in\R.
\ee
Let us assume this claim for a moment and proceed to prove the theorem.
Because of (i) in Lemma \ref{L:Rhat},  we know that each of the functions $\hat R_1^n,\hat R_2^n$ is in $\Upsilon^{5,n}(\Relu;1,1)$.
Since $g$ is a CPwL function with $m$ breakpoints, we have $g\in \Upsilon^{m,1}(\Relu;1,1)$.  We can output $g(\hat R_1^n)$ by concatenating the networks for $\hat R_1^n$ with that for  $g$.  Namely, we place the neural network for $\hat R_1^n$ in the first $n$ layers and then follow that with the network for  $g$ in the last layer using $\hat R_1^n(x)$ as its input.  Thus, $g(\hat R_1^n)$ is in $\Upsilon^{\max\{m,5\},n+1}(\Relu;1,1)$.  We now place the two neural networks that output
$g(\hat R_1^n)$ and $g(\hat R_2^n)$  stacked on top of each other.  The resulting  network has width $W=2\max\{m,5\}$.  At the final step, we recall that
\be
\label{min}
\min\{x,y\}= y_+-(-y)_+ -(y-x)_+.
\ee
Therefore, by  adding another layer to  the network we have already created, we can output the right side of \eref{claim11}. Hence, it 
is in $\Upsilon^{W,n+2}(\Relu;1,1)$ with the advertised width $W$.

\begin{figure}
	\centering
	\begin{subfigure}[b]{0.475\textwidth}
		\centering
		\includegraphics[width=\textwidth]{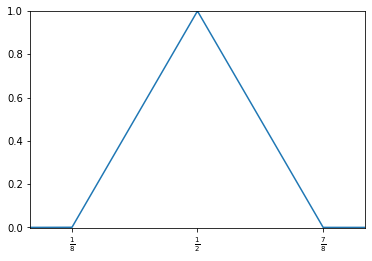}
		\caption{Graph of the special function $g$.}
		
	\end{subfigure}
	\hfill
	\begin{subfigure}[b]{0.475\textwidth}
		\centering
		\includegraphics[width=\textwidth]{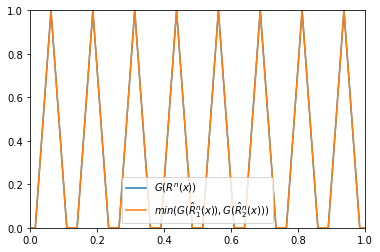}
		\caption{Graphs of  $\min\{g(\hat{R}^n_{\alpha_1,\beta_1}), g(\hat{R}^n_{\alpha_2,\beta_2})\}$ and $g(R^n)$ on $[0,1]$}
		
	\end{subfigure}

	\caption{Comparison between  $g(R^n)$ and $\min\{g(\hat R^n_{\alpha_1,\beta_1}), g(\hat R^n_{\alpha_2,\beta_2})\}$
		on $[0,1]$, where  $(\alpha_1,\beta_1,\alpha_2,\beta_2) = (0.492,0.494,0.496,0.498)$. 
	}
	\label{fig:G_plot}
\end{figure}

With these remarks in hand, we are left with proving the claim.  For the proof, let us  first note that \eref{claim11} holds when $x$ is outside $[0,1]$ because   $R^n(x)=\hat R_1^n(x)=\hat R_2^n(x)$ outside $[0,1]$.  Now, consider a general $\hat R=\hat R_{\alpha,\beta}$ and understand what $g(\hat R^n(x))$   looks like as $x$ traverses a dyadic interval
$I_k:=[k2^{-n},(k+1)2^{-n}]$ from left to right. We will track this behavior only when $\alpha$ is extremely close to
$1/2$, i.e. $\delta=1/2-\alpha $ is very small; at least $0<\delta\ll2^{-n}$.
From (ii) of Lemma \ref{L:Rhat}, we know that $\hat R^n(x)=R^n(x)$  until $x$ gets close to the right endpoint of $I_k$, and in particular on 
$$
I_k\setminus J_k(\alpha)\subset \Omega_n,  
\hbox{where  } J_k(\alpha):= [(k+1)2^{-n}-\delta,(k+1)2^{-n}].
$$
This means that $g(\hat R^n(x))=g(R^n(x))$   until $x\in J_k(\alpha)$.
On $J_k(\alpha)$, we have $g\circ \hat R^n\ge 0$ since $g$ is a special function (and therefore non-negative).  By (iii) of Lemma \ref{L:Rhat}, $\hat R^n(x)=0$  on  
$$
J_k'(\beta):= [(k+1)2^{-n}-\delta' 2^{-n+1},(k+1)2^{-n}]\subset J_k(\alpha)
,$$
where
$$
\delta':=1/2-\beta,
$$
and therefore $g\circ \hat R^n\equiv 0$
on $J_k'(\beta)$.

Now, we return to  our choice of $ \alpha_1,\beta_1,\alpha_2,\beta_2$.    We first choose the  $\alpha_1<\beta_1$ very close to $1/2$ so that at least
$\delta_1:=1/2-\alpha_1<2^{-n-3}$, which ensures that 
$$g\circ R^n\equiv 0\quad \hbox{on}\quad  J_k(\alpha_1):=[(k+1)2^{-n}-\delta_1,(k+1)2^{-n}]
$$
because of \eref{know10}. Moreover,
it follows from the discussion so far that 
$$
g\circ R^n=g\circ \hat R_1^n\quad \hbox{on}\quad  I_k\setminus J_k(\alpha_1).
$$
For any $ \beta_1<\alpha_2<\beta_2<1/2$
we have that $g\circ R^n=g\circ \hat R_1^n=g\circ \hat R_2^n$ on $I_k\setminus J_k(\alpha_1)$, but 
we choose $\alpha_2$ so that 
$$
\delta_2:=1/2-\alpha_2< \delta_1'2^{-n +1}:=(1/2-\beta_1)2^{-n +1}.
$$
This guarantees that
for $x\in J_k(\alpha_1)$ one of the two non-negative numbers $g(\hat R_1^n(x))$ or
$g(\hat R_2^n(x))$ is zero, and thus we have \eref{claim11}.
\hfill $\Box$


\subsection{Matrix constructions}

In this section, we continue considering 
special functions $g$ and their 
vectorization $G$.  
Let us introduce the notation
$$
\M_j(x):=T_{B_j(x)},\quad x\in[0,1], \quad j=1,2,\dots,
$$
for the piecewise constant matrix valued
functions $T_{B_j(x)}$.  
Let $T(x)$ be the piecewise constant matrix valued function which is defined as 
\begin{eqnarray}
	\label{Tdef}
	T(x) &:=&
	\chi_{[0,1/2)}(x) \cdot T_0 + \chi_{[1/2,1]}(x) \cdot T_1\\
	&=& \left\lbrace \begin{array}{ll}
		T_0, & x <1/2, \\
		T_1, & x  \geq 1/2.
	\end{array}\right.
\end{eqnarray}
Notice that we have purposefully defined $T$ on all of $\R$.  We have
\be
\label{Mk}
\M_j(x)= T(R^{j-1}(x)),\quad x\in [0,1], \quad j=1,\ldots,
\ee
if we set
$$
R^0(x):=x, \quad x\in [0,1].
$$
We know from the cascade algorithm \eref{cascade} that 
\be
\label{know}
G_n(x)= \M_1(x)\ldots\M_n(x)G(R^n(x)), \quad x\in[0,1].
\ee

We next  introduce our technique for proving that $G_n$ is an output of a  neural network whose depth grows linearly in $n$. 
In what follows, we derive a new expression for $G_n$ and then use it to prove the existence of such a neural network. Recall that for a special function $g$,  
$$
G(x)=g(x)e_1, \quad x\in[0,1],
$$
where the column vectors $e_j\in \R^N$, $j=1,\ldots,N$, are the standard basis for $\R^N$. Then \eref{know} can be rewritten as
\begin{eqnarray}
	G_n(x)&=&(\bar g_1(x),\ldots,\bar g_n(x))^T \nonumber\\
	&=&\M_1(x)\ldots\M_n(x)
	g(R^n(x))e_1,\nonumber
\end{eqnarray}
for $x\in[0,1]$,  
and thus the $k$-th coordinate $\bar g_k$ of $G_n$  is given by
\begin{equation}\label{eq:gk}
	\bar g_k(x)=g(R^n(x)) e_k^T\M_1(x)\cdot \ldots \cdot \M_n(x)e_1.
\end{equation}

Recall that since $g$ is special,  see \eref{know10}, we have 
\begin{eqnarray}
	&g(R^n(x)) = 0, \nonumber\\ &x\in \Lambda:= [0,1]\cap \bigcup_{0\le j\le 2^n} [j2^{-n}-2^{-n-3},j2^{-n}+2^{-n-3}].\nonumber
\end{eqnarray}

Because of this, for each $j$, we can modify the definition of $\cM_j$ to $\hat \cM_j$  as long as $\hat \cM_j$ agrees with $\cM_j$  
on $\Lambda':=[0,1]\setminus \Lambda$ and  formula \eref{eq:gk} will
remain valid.  We now describe the modification we shall use.

It follows from \eref{Mk} and \eref{Tdef} that for $j=1,\ldots,n$, 
\begin{eqnarray}
	\label{w2}
	\M_j(x) &=& T(R^{j-1}(x))\nonumber\\
	&=&  \chi_{[0,1/2{ )}}(R^{j-1}(x)) \cdot T_0 + \chi_{{ [}1/2,1]}( R^{j-1}(x)) \cdot T_1
	,\quad x\in [0,1],
\end{eqnarray} 

where $R^0(x):=x$ and  $\chi_I$ is the  indicator function of the interval $I$.   
Note that the functions $\chi$ and $R^{j-1}$, $j\geq 2$, participating in \eref{w2} are discontinuous.   We replace the characteristic functions 
$\chi_{[0,1/2{ )}}$ and $\chi_{{ [}1/2,1]}$
by the CPwL functions
$$
\hat\Chi_{0}(x) := 
1  - \frac{1}{\delta_0}(x - 1/2)_{+}  + \frac{1}{\delta_0}(x - 1/2 -\delta_0)_{+},
$$
$$
\hat\Chi_{1}(x) :=-\frac{1}{\delta_0}(x - 1/2 )_{+} + \frac{1}{\delta_0}(x - 1/2 + \delta_0)_{+},
$$
respectively, where $\delta_0:=2^{-n-3}$.   These functions satisfy
for $x \notin [1/2-2^{-n-3}, 1/2+2^{-n-3}]$
\begin{eqnarray}
	\label{satisfy}
	\hat\Chi_0(x)&=& \chi_{[0, 1/2)}(x) \text{ and }\nonumber\\  \hat\Chi_1(x)&=&\chi_{ [1/2,1]}(x).
\end{eqnarray} 

For $R$, we can use one of the substitutes $\hat R_{\alpha,\beta}$ defined in the previous section. We take  $\alpha>0$ such that  $\delta:=1/2-\alpha<2^{-n-3}$, $\beta\in (\alpha,1/2)$ and then  consider
$$
\hat R:= \hat R_{\alpha,\beta}. 
$$ 
With these preparations in hand, we define for $x\in [0,1],\ j=1,\dots,n,$
\begin{equation}
	\label{hatMj}
	\hat \cM_j(x):=  \hat\Chi_0(\hat R^{j-1}(x))T_0  + \hat\Chi_1( \hat R^{j-1}(x))T_1,
\end{equation}
where $\hat R^0(x):=x=R^0(x)$.

Finally, let us recall the sets $E_j$, $j=1,\dots,n$, see  \eref{defEj}, and define
$$
E:=\bigcup_{j=1}^n E_j, \quad  E':=  [0,1]\setminus E.
$$
Note that $E'$ was denoted by $\Omega_n$ earlier.
\begin{remark} 
	\label{R:important}
	Notice that $E\subset \Lambda$ 
	because of the choice
	of $\delta<2^{-n-3}$, and so
	$\Lambda' \subset E'$.  This means that for each $j= 1,\dots,n-1$, we have $\hat R^j(x)=R^j(x)$ for  $x\in \Lambda'$,  see {\rm (ii)} of Lemma {\rm \ref{L:Rhat}}.
\end{remark}

\begin{lemma} 
	\label{L:equality}
	For each $j=1,2,\dots,n$,  we have 
	\be 
	\label{equal}
	\hat\cM_j(x):=\cM_j(x),\quad x\in \Lambda',
	\ee 
	and therefore \eref{eq:gk} holds with each $\cM_j$ replaced by $\hat\cM_j$, $j=1,2,\dots,n$.
\end{lemma}
\noindent{\bf Proof:}  Let $x\in \Lambda'$ and $1\le j\le n$. From  Remark \ref{R:important},  we know that
$\hat R^{j-1}(x)= R^{j-1}(x)$.
Next, we claim that when $x\in \Lambda'$, we have for $j=0,\dots,n-1$,
\begin{equation}
	\label{claim5}
	R^j(x)\notin { [}1/2-2^{-n-3}, 1/2+2^{-n-3}]=:I \subset \Lambda,
\end{equation} 
which in view of \eref{satisfy} will complete the proof of the Lemma. 

To prove the claim \eref{claim5}, let us first note that when $j=0$ the claim
is true because $R^0(x)=x$ and $\Lambda'$ does not contain points from  $I$.   Now consider $j\ge 1$ and let $x\in  \Lambda'$.  Choose $l$ so that
$x\in [l2^{-j},(l+1)2^{-j})=:J$. 
Recall that
$R^{j}(x)=2^{j}x-l$, for $x\in J$, and hence this function takes the value $1/2$ at $2^{-j}(l+1/2)$.  Since $x\in \Lambda'$, it is at distance
at least $\delta_0=2^{-n-3}$ from $(l+1/2)2^{-j}$. Therefore, $R^j(x)$
differs from $1/2$ by at least $2^j\delta_0$.  This means $R^j(x)$ is not in $I$ as desired. \hfill $\Box$

We conclude this section with the observation that we now know that
the $k$-th coordinate $\bar g_k$ of $G_n$
has the representation
\begin{equation}
	\label{eq:gk1}
	\bar g_k(x)=g(R^n(x)) e_k^T\hat \M_1(x)\cdot \ldots \cdot \hat\M_n(x)e_1,
\end{equation}
whenever  $g$ is a special function.

\subsection{Computing products}
\label{S:products}
In the next section, we use \eref{eq:gk1}  to  show that $\bar g_k$ is the output of an NN with fixed width and depth $Cn$ with $C$ a fixed constant. If we look at \eref{eq:gk1}  and think of implementing it from left to right, we will be required to compute products of the form $aA$,  with  $a$ being a row vector and $A$ being a matrix, and both being outputs of 
an NN. 
In general,  such a product is not an  output of an NN.  In our case, however, both $a,A$ have special properties which will allow us to show that the products we need are actually outputs of an NN.   
The following lemma describes one setting when NNs can output products exactly. We describe the simplest case  where we want to take the product of a scalar valued function with
a vector valued function.   More precisely, we propose a function $\Pi=\Pi(x,y)$ that is an output of an NN and matches the true product $xy$  whenever the scalar has values $0$ or $1$ or the vector $y$ is the zero vector.   This will match our needs for computing $\bar g_k$.

In what follows, we denote by 
${\bf e}\in \R^N$ the column vector whose coordinates are all ones, that is
$$
{\bf e}:=(1,1,\ldots,1)^T.
$$
\begin{lemma}
	\label{L:Shahar}
	Let $M$ be a positive number and let us define
	the function $\Pi:\R \times\R^{N}\to \R^N$, by
	\begin{eqnarray}
		\label{eq:rho}
		\Pi(x,y):=&-&\relu(Mx{\bf e}-y)\nonumber \\ 
		&-& \relu\left(M(1-x){\bf e}-\relu(-y)\right)\nonumber\\
		&+&M{\bf e},
	\end{eqnarray}
	for $x\in \R$ and $ y\in \R^N$.
	Then, for all  $y=(y_1,\ldots,y_N)^T$ with $|y_i|\leq M$, and all $x\in[0,1]$, we have  
	\begin{eqnarray}
		\label{N1}
		&\Pi \in\Upsilon^{2N+1,2}(\Relu;N+1,N),\nonumber \\ 
		&\Pi(1,y)=y,\nonumber \\ 
		&\Pi(0,y)=0, \nonumber\\
		&\Pi(x,0)=0.
	\end{eqnarray}
\end{lemma}
{\bf Proof:} We consider  an NN with $N+1$ inputs $(x,y)\in \R^{N+1}$. Its
first hidden layer has $2N+1$ nodes, holding 
$\relu(Mx{\bf e}-y)$, $\relu(-y)$ and $x=\Relu(x)$. Its second layer has $2N$ nodes holding the values
$\relu(Mx{\bf e}-y)$ and  
\begin{multline}
	$$\relu\left(M(1-x){\bf e}-\relu(-y)\right)=\\
	\relu\left(-M\Relu(x){\bf e}-\relu(-y)+M{\bf e}\right).
	$$
\end{multline}
We can view the first $N$ nodes in the second layer either as ReLU-free nodes, or we can add a large bias which we then subtract after re-weighting at the output.
Finally, its output layer has $N$ nodes and bias $M{\bf e}$.  Clearly, this network outputs $\Pi$ in the case when $x\in[0,1]$.

A direct computation shows that
\begin{eqnarray}
	\Pi(1,y)&=&- \relu(M{\bf e}-y) - \relu\left(-\relu(-y)\right)+M{\bf e}\nonumber\\
	&=&-M{\bf e}+y+M{\bf e}\nonumber\\
	&=&y,\nonumber
\end{eqnarray}
since $|y_i|\leq M$, $i=1,\ldots,N$, and $\relu\left(-\relu(-y)\right)=0$. Likewise,

\begin{eqnarray}
	\Pi(0,y)&=&- \relu(-y) - \relu\left(M{\bf e}-\relu(-y)\right)+M{\bf e}\nonumber\\
	&=&
	- \relu(-y) -M{\bf e}+\relu(-y)+M{\bf e}\nonumber\\
	&=&0,\nonumber 
\end{eqnarray}
and
\begin{eqnarray}
	\Pi(x,0)&=&-\Relu(Mx{\bf e})-\Relu(M(1-x){\bf e})+M{\bf e}\nonumber\\
	&=&-Mx{\bf e}-
	M(1-x){\bf e}+M{\bf e}\nonumber\\
	&=&0,\nonumber
\end{eqnarray}   
since $|y_i|\leq M$, $i=1,\ldots,N$, and $x\in [0,1]$.     
\hfill $\Box$

\subsection {The coordinates of $G_n$ are  outputs of an NN}
In this section, we show that each coordinate $\bar g_k$ of $G_n$ is the output of an NN
with fixed width $W$ and depth $Cn$ with $C$ a fixed constant.  We let $k\in\{1,\dots,N\}$
be arbitrary but fixed in this section.  We begin with the representation
\eref{eq:gk1} for $\bar g_k$ and implement the product, from left to right,  via a recursive procedure.
Namely,  we
recursively define for $\\ j=1,\ldots,n$,
\begin{eqnarray}
	F^0(x)&=&g(R^n(x)) e_k^T, \nonumber\\
	F^j(x)&=&F^{j-1}(x){\hat \M}_j(x).\nonumber
\end{eqnarray}
So, each $F^j(x)$ is  a row vector.  Clearly, we have
\begin{equation}\label{eq:Gnk}
	\bar g_k(x)=F^n(x)e_1.
\end{equation}
We now replace $\hat \cM_j$ by its representation \eref{hatMj}, write everything as column vectors and obtain the recursion for $j=1,\ldots,n,$
\begin{eqnarray}
	\label{pq}
	[F^j(x)]^T&=&\hat\Chi_0(\hat R^{j-1}(x))\cdot \cF^{j-1}_0(x) +\nonumber\\
	&&\hat\Chi_1(\hat R^{j-1}(x))\cdot \cF^{j-1}_1(x),
\end{eqnarray}
with 
\be
\label{pq1}
\cF^{j-1}_0(x)=T_0^T[F^{j-1}(x)]^T,\quad 
\cF^{j-1}_1(x)=T_1^T[F^{j-1}(x)]^T.
\ee
Notice that each of the two terms appearing in \eref{pq} is the product of a scalar times
a vector and so  we can use $\Pi$ to implement this product.
For this, we define
$$
M:=\max_{j=0,\ldots,n-1}\max_{x\in[0,1]}\left\{
\|\cF_0^{j}(x)\|_{\ell_\infty^N},
\|\cF_1^{j}(x)\|_{\ell_\infty^N}\right\},
$$
and define  the function $\Pi(\cdot,\cdot)$ in \eref{eq:rho} with this constant $M$.

\begin{lemma}
	\label{L:Pi}
	For each $j=1,\dots,n$, we have
	\begin{eqnarray}
		\label{w21}
		[F^j(x)]^T &=&   \Pi(\hat\Chi_0(\hat R^{j-1}(x)),\cF_0^{j-1}(x))   +\nonumber \\ 
		&&\Pi(\hat\Chi_1( \hat R^{j-1}(x)),\cF_1^{j-1}(x)).
	\end{eqnarray}
\end{lemma}
\noindent
{\bf Proof:}  This is proved by induction on $j$.   Consider first the case $j=1$.
Since 
$g(R^n)$ vanishes on $\Lambda$, we have that $F^0=g(R^n)e_k^T$,
$\cF^0=T_0^T[F^0]^T$, and
$\cF^1=T_1^T[F^0]^T$
vanish on $\Lambda$. Moreover, since
both  $\hat \Chi_0(\hat  R^0)$ and 
$\hat\Chi_1(\hat R^0)$ take only the values $0$ or $1$ outside of $\Lambda$,  Lemma \ref{L:Shahar} gives the result.  We also  obtain  that $F^1$ vanishes on $\Lambda$.

For $j\ge 2$, we proceed by induction, using that $F^{j-1}$ vanishes on $\Lambda$ and relations  \eref{claim5} and \eref{satisfy} for $x\in \Lambda'$.  Namely, we use the same argument to
prove \eref{w21} and also to prove that $F^j$ vanishes on $\Lambda$.  This allows us to advance the induction.\hfill $\Box$

\begin{theorem}
	\label{MainT}
	If $g$ is a special function with $m$ break points, then for each $k\in\{1,\dots,N\}$ and each $n\geq 1$, the coordinate function $\bar g_k$ of $G_n$ belongs to the set
	$\Upsilon^{W,L}(\Relu;1,1)$, where 
	$$
	W=\max\{1+2\max\{m,5\},N+10,4N+3\},
	\quad 
	L=4n+1.
	$$
\end{theorem}
\noindent
{\bf Proof:}
We fix $k$ and construct a neural network $\cN$  of the advertised size that outputs $\bar g_k$.  The network $\cN$ is the concatenation of several networks: a network $\cN_0$ and then $n$ copies of a network $\cN'$.  Let us note before we begin the proof that we repeatedly use Remark \ref{R:fill} 
so that all of the component networks have the same width. We do not always elaborate on that aspect going further. 

The first network $\cN_0$ has input $x$ and outputs $(x,g(R^n(x)))=(\hat R^0(x),g(R^n(x)))$.
It  consists of  the network producing $g(R^n)$  augmented    with an additional channel that passes forward the value $x$.
We can view this channel as a ReLU channel since $x\geq 0$.
From 
Theorem \ref{T1} we know that 
\be
\label{NNg}
g\circ R^n\in\Upsilon^{2\max\{m,5\},n+1}
(\Relu;1, 1),
\ee
where $m$ is the number of breakpoints of $g$. 
Thus, the width of this network is $W_0=1+2\max\{m,5\}$ and the depth is $L_0=n+1$.

We come now to the recursive part $\cN'$ of $\cN$, where  $\cN$ is completed by concatenating  $\cN_0$ with $n$ copies of
$\cN'$.
For each $j=1,\dots, n$,  the network $ \cN'$ that we describe below   will take as
\vskip .1in
\noindent{\bf Input:} $  \hat R^{j-1}(x),F^{j-1}(x),$ 
\vskip .1in
\noindent
and will produce  
\vskip .1in
\noindent
{\bf Output:}  $\hat R^{j}(x), F^{j}(x)$.
\vskip .1in
\noindent
Notice that $\cN_0$ can output $\hat R^0(x)=x$ and $F^0(x)=g(R^n(x))e_k^T$ to start the iteration.
After these 
$n$ iterations are completed, $\cN$ can output $F^n(x)$ and $\bar g_k(x)$.  Thus, it is enough to show that each iteration can be completed with an NN of fixed width and depth. 

We now describe the network $\cN'$ 
that gives the
{\bf Input} to {\bf Output} above.  It has depth $L'=3$. The first node of the first layer forwards the value $\hat R^{j-1}(x)$. The next $5$  nodes of this layer are the network for $\hat R$, followed by the $2$ nodes of the network for $\hat\Chi_0$ and the $2$ nodes for the network for $\hat\Chi_1$. The last $N$ nodes hold $F^{j-1}(x)$. 
Thus, the width of this layer is $N+10$.  

The next two layers of $\cN'$  consist of two copies of the network producing $\Pi$ on the top of each other, augmented with additional  channel that holds $\hat R^j(x)$. Thus, the width of these two layers is $4N+3$. So, the width of $\cN'$
is $W'=\max\{N+10,4N+3\}$.
Clearly, then the width of $\cN$ is
$$
W=\max\{1+2\max\{m,5\},N+10,4N+3\},
$$
and the
depth is
$$
L=L_0+nL'=n+1+3n=4n+1.
$$

We now describe how these networks are concatenated to produce $\cN$. We feed the output $\hat R^0(x)$ into the first node of the first layer and into the networks for $\hat R$, $\hat \Chi_0$ and $\hat\Chi_1$. We feed zeroes to the last $N$ nodes of this layer, except to the $k$-th of those, where we feed $g(R^n(x))$. Thus, the output of the first layer of $\cN'$ is 
$(\hat R^0(x),\hat R^1(x),\hat\Chi_0(\hat R^0(x)),\hat\Chi_1(\hat R^0(x)),F^0(x))$. 

We then keep $\hat R^1(x)$ by feeding it through the first node of the remaining layers before outputting it. We use the entries of the matrix $T_0$ as weights applied to the output $F^0(x)$ of the previous layer to produce $\cF^1_0(x)$ and feed  it to the first copy of the network for $\Pi$. We feed to this first copy the output $\hat\Chi_0(\hat R^0(x))$ of the previous layer. We use the entries of the matrix $T_1$ as weights applied to the output $F^0(x)$ of the previous layer to produce $\cF^1_1(x)$ and feed it, together with the output $\hat\Chi_1(\hat R^0(x))$ from the previous layer, to the second copy of the network for $\Pi$.

We use formula \eref{w21} to output $F^1(x)$. Thus, after the concatenation of $\cN_0$ and $\cN'$, we have computed $(\hat R^1(x),F^1(x))$. Then, we concatenate as discussed above with the next copy of $\cN'$ and so on.\hfill $\Box$

\subsection{Piecing together}
\label{S:piecing}

In this section, we construct an NN of controllable size that outputs
$V^ng$ whenever $g$ is a special function.  This is done by using  the NNs from Theorem \ref{MainT} that produce the components $\bar g_k$, $k=1,\dots,N$, of  $G_n$.
\begin{theorem}
	\label{MainT1}
	If $g$ is a special function
	with $m\geq 3$ breakpoints, then for each $n\geq 1$,
	\begin{multline}
		\label{pop}
		V^ng\in\Upsilon^{NW,4n+2}(\Relu;1,1),\quad \\ \hbox{where}\quad  W=\max\{1+2\max\{m,5\},N+10,4N+3\}.
	\end{multline}
\end{theorem}
\noindent
{\bf Proof:}
We  introduce the  following `ramp' functions $r_k\in \Upsilon^{2,1}(\Relu;1,1)$, $k=1,2,\ldots,N$,

\begin{eqnarray}
	\label{VGwrite}
	r_k(x)&=&(x-k+1)_+-(x-k)_+\nonumber\\
	&=&
	\begin{cases}
		0,\quad x\leq k-1,\\
		x-k+1,\quad k-1\leq x\leq k,\\
		1,\quad x\geq k.
	\end{cases}
\end{eqnarray}

It follows  that 
\begin{multline}
	\label{imp}
	r_k(x)=x-k+1\quad \hbox{when}\quad x\in [k-1,k],
	\quad \hbox{and} \quad \\ r_k(x)\in\{0,1\}\quad 
	\hbox{when}\quad x\in \R\setminus [k-1,k].
\end{multline}
We use these functions to  construct an NN $\cN_1$   with input $x\in\R$ and output   
$(r_1(x),\ldots,r_N(x))\in\R^N$ by stacking on the top of each other the networks for $r_1,r_2, \ldots,r_N$.  This network 
has width $2N$ and depth $1$.
Next, we build a network $\cN_2$ by stacking on the top of each other the NNs from Theorem \ref{MainT} that produce $\bar g_k(x)$, $k=1,\ldots,N$. This network has width $NW$, where $W$ is the width from Theorem \ref{MainT} and depth $4n+1$. We concatenate $\cN_1$ with $\cN_2$ by feeding the first coordinate $r_1(x)$ of the output of $\cN_1$ to the NN producing $\bar g_1$, 
the second coordinate $r_2(x)$ of the output of $\cN_1$ to the NN producing $\bar g_2$, etc. 
The concatenated neural network  has width $NW$, depth $4n+2$ and outputs 
$$
\bar g_1(r_1(x))+ \bar g_2(r_2(x))+\ldots+ \bar g_N(r_N(x)).
$$
Moreover, because of property 
\eref{imp} and the fact that 
$\bar g_k(0)= \bar g_k(1)=0$ for all $k=1,\ldots,N$, the output of the constructed NN will be exactly  $V^ng(x)$.
\hfill $\Box$

\begin{remark}
	Other strategies to construct an NN that outputs  $V^ng$ using the pieces $\bar g_k$, $k=1,\ldots,N$, are possible. We have discussed a strategy that inflates the width of the resulting NN (thus obtaining a width that is possibly $\asymp N^2$), but keeps the depth $L$ independent of $N$. One may choose to inflate the depth by a factor of $N$
	(thus having depth $L\asymp Nn$)
	but keep the width possibly linear in $N$.
\end{remark}

\subsection{Proof of Theorem \ref{T:maintheorem}}
We first wish to start the discussion with $g=g_0$ a CPwL function that has support in $[0,N]$ and has breakpoints at the integers. 
\vskip .1in
\noindent
{\bf Proof of Theorem \ref{T:maintheorem} for $g_0$:} 
Recall that  \eref{newform}  gives $V^n g_0$ as a linear combination of shifts of $V^nH$, where $H$ is the special function, defined in \eref{gg}, that is 
$$
V^n g_0(x)=
\sum_{j=1}^{8N-1}
g_0(j/8)V^nH(x-2^{-n-3}(j-4)).  
$$
Since $H$ has $m=3$ breakpoints, it follows from Theorem \ref{MainT1} that 
$$
V^nH\in \Upsilon^{N\max\{N+10,4N+3\},4n+2}(\Relu;1,1).
$$
We now use two basic well-known facts about outputs of NNs.
\begin{itemize}
	\item If the function $S$ belongs to $\Upsilon^{W',L'}(\Relu;1,1)$, then for any shift $\delta$, the function $S(\cdot +\delta)$ is in
	$ \Upsilon^{W',L'}(\Relu;1,1)$.
	\item If $S_1,\dots,S_\ell\in \Upsilon^{W',L'}(\Relu;1,1)$, then the sum $\sum_{j=1}^\ell S_j\in\Upsilon^{W'+2,\ell L'}(\Relu;1,1)$.
\end{itemize}
Therefore, for every $j=1,\ldots,8N-1$,
\begin{eqnarray}
	& V^nH(\cdot-2^{-n-3}(j-4))\in \nonumber\\ &\Upsilon^{N\max\{N+10,4N+3\},4n+2}(\Relu;1,1),
\end{eqnarray}
and
$$
V^ng_0\in \Upsilon^{N\max\{N+10,4N+3\}+2,
	(8N-1)(4n+2)}(\Relu;1,1).
$$
The proof is completed.
\hfill $ \Box$

Let us now explain why the theorem holds for any CPwL function $g$ supported on $[0,N]$, i.e., why it is not   necessary to assume that the breakpoints of $g=g_0$ are at the integers.  If $g$ has arbitrary breakpoints, we can always take  a finite set of breakpoints which contain the original breakpoints of $g$ and represent $g$ as a linear
combination of hat functions centered at these new breakpoints.  By choosing the new breakpoints fine enough, each of the   hat functions will have a shift which is a special function and so the above proof can be carried out in the same way.  Thus, any CPwL function $g$ supported on $[0,N]$
has the property that $V^ng\in \Sigma_n$. 

\section{Approximation of refinable functions}
\label{S:properties}

In this section,  we show how the results of the previous section give
upper bounds on how well refinable functions can be approximated by deep NNs.  This is accomplished by utilizing  known results from the theory of refinable functions. Under certain conditions on the refinement mask $\bar c$, the refinement equation has a unique solution $\phi=\phi_{\bar c}$ (up to scaling). Moreover, for certain
initial choices of a CPwL function 
$\phi_0$, the cascade algorithm gives that $V^n\phi_0$ converges to $\phi$
in $L_\infty$ with an exponential  decay rate.  There is a variety of statements of this type, and rather than trying to summarize all of them, we refer the reader to the existing literature, especially \cite{DL,DL1} and the references in those papers. 

We say a refinement mask $\bar c$  is {\it admissible} if there exists a CPwL
function 
$\phi_0$ supported on $[0,N]$ for which
\be 
\label{admissible}
\|\phi_{\bar c}-V^n \phi_0\|_{L_\infty(\R)} \le C\lambda^n,\quad n\ge 1,
\ee 
where $0<\lambda<1$ and $C>0$ are   constants depending only on $\bar c$.  Since the preceding section shows
that $V^n\phi_0\in\Sigma_n:=\Upsilon^{C',Cn}(\Relu;1,1)$, with $C',C$ depending only on $\bar c$ and the number of breakpoints $m$ of 
$\phi_0$, we obtain the following theorem

\begin{theorem}
	\label{T:approxphi}
	If $\bar c$ is admissible then for $\Sigma_n$ as defined immediately above, we have
	\be
	\label{phierror}
	E_n(\phi)_{L_\infty(\R)} :=\inf_{S\in\Sigma_n}\|f-S\|_{L_\infty(\R)} \le  C\lambda^n,\quad n\ge 1.
	\ee 
\end{theorem}

There are many results that  establish conditions on $\bar c$ that guarantee admissibility.   In one set of results, one takes for 
$\phi_0$ the CPwL function that
interpolates $\phi_{\bar c}$ at the integers $\{0,1,\dots,N\}$ and has its only breakpoints
at these integers.  A second set of results shows that under certain conditions on
$\bar c$, one can take for 
$\phi_0$  the hat function centered at one with base width two.

\section{Relation to $n$-term wavelet approximation and subdivision}
\label{S:NNA}
We have shown that univariate refinable functions $\phi$ can be approximated very efficiently using ReLU neural networks.  In this section,
we wish to point out some consequences of this result related to $n$-term approximation from a dictionary and to subdivision algorithms.

Let us first continue with a discussion of the approximation of functions of one variable.  One common tool for approximation, prevalent
in computational harmonic analysis, is to use wavelets and their descendants in the approximation.  The classical  wavelets $\psi$ such as the Daubechies
compactly supported wavelets (see \cite{D10}) or the Cohen-Daubechies-Feauveau  biorthogonal wavelets (see \cite{CDF}) are each a finite linear combination of shifts of a refinable function $\phi$.  Therefore, it follows from our results that $\psi$ is  very well approximated by outputs of neural networks.  Namely,
the bounds on approximation error that we have derived for $\phi$ apply for $\psi$ as well.  This can then be used to approximate general functions
$f$ via their wavelet decomposition.  We briefly discuss  how this plays out when the approximation takes place in $L_2 (\R)$.  A similar analysis holds when approximation takes place in $L_p(\Omega)$ with $\Omega$ a sub-interval of $\R$.

We suppose that $\psi$ is a compactly supported wavelet which generates an orthogonal basis. To describe this wavelet basis, we normalize $\psi$ so that   $\|\psi\|_{L_2(\R)}=1$. Let $\cD$ be the set of dyadic intervals $I=[j,j+1]2^{-k}$ where $j,k$ are integers.  Then the functions 
\be 
\label{wavelets}
\psi_I(x)=2^{k/2}\psi (2^kx-j),\quad I=[j,j+1]2^{-k},
\ee
form a complete orthonormal system $\Psi:=(\psi_I)_{I\in\cD}$ for $L_2(\R)$.
Therefore, every  function $f\in L_2(\R)$
has a wavelet decomposition
\be
\label{wd}
f=\sum_{I\in \cD} f_I \psi_I,\quad f_I:=\langle f,\psi_I\rangle{:= \int_{\R}f\psi_I}.
\ee

A common approximation and numerical tool is to use an $n$-term approximation to
$f$ from the terms in \eref{wd}.  This type of approximation has error
\be
\label{ntermerror}
\sigma_n(f)_{ L_2(\R)}:=\inf_{\#(\Lambda)=n} \|f-\sum_{I\in\Lambda}f_I\psi_I\|_{L_2(\R)}.
\ee
Because of the results we have proved about NNA of $\psi$,  we can replace
each $\psi_I$ by $\hat \psi_I$ from  $\Upsilon^{C',C\ell}(\Relu;1,1)$ and incur an additional error in the approximation for each term that is replaced.  Consider, for example, the case when $\psi$ is approximated to accuracy $C2^{-\ell\alpha}$ by the elements of $\Sigma_\ell:=\Upsilon^{C',C\ell}(\Relu;1,1)$.   Then, each function $\psi_I$ is likewise approximated by a $\hat \psi_I\in\Sigma_\ell$  with the same accuracy $C2^{-\ell\alpha}$.  This gives that any sum $S=\sum_{I\in\Lambda}f_I\psi_I$ of $n$ terms is approximated by $\hat S:=\sum_{I\in\Lambda}f_I\hat \psi_I$  with accuracy
\be 
\label{nerror}
\|S-\hat S\|_{L_2(\R)}\le  C2^{-\ell\alpha}\sum_{I\in\Lambda} |f_I| \le C\sqrt{n}2^{-\ell\alpha}\|f\|_{L_2(\R)},\quad n\ge 1. 
\ee 
Since $\hat S$ is in $\Sigma_{Cn\ell}$, this gives the approximation rate   %
\be
\label{errorn}
E_{Cn\ell}(f)_{L_2(\R)}\le C\sigma_n(f)_{L_2(\R)}+C\sqrt{n}2^{-\ell\alpha}.
\ee

Let us consider an example, where $\sigma_n(f)_{L_2(\R)}\le Mn^{-r}$ with $r>0$. Choosing $\ell= a\log_2 n$ with $a$ sufficiently large depending only on $r$, we obtain 
\be 
\label{obtain}
E_{n\log_2 n}(f)_{L_2(\R)}\le Cn^{-r},\quad n\ge 1.
\ee 
In other words, using slightly more parameters in NNA achieves the approximation rate
of $n$-term wavelet approximation. This rate holds for all $r>0$ and universally for all
choices of the wavelet basis.  It does not require knowledge of the wavelet $\psi$.

The above results extend to $n$-term wavelet approximation of multivariate functions defined on $\R^d$ or a domain $\Omega\subset \R^d$.
The standard construction of multivariate wavelet bases uses shifted dilates of functions 
\be
\label{mvwave}
\psi^{e}(x_1,\dots,x_d) =\psi^{e_1}(x_1)\cdots \psi^{e_d}(x_d), \quad e\in \{0,1\}^d,
\ee
where $ \psi^{(0)}:=\phi$ and  $\psi^{(1)}:=\psi$.  The wavelet decomposition of
$f\in L_2(\Omega)$ now takes the form
\be
\label{wd1}
f=\sum_{I\in \cD(\Omega)} f_{I,e} \psi^e_I,\quad f_{I,e}:=\langle f,\psi^e_I\rangle,
\ee
where now $\cD(\Omega)$ is a collection of dyadic cubes supported near
$\Omega$ and the $\psi_I^e$ are scaled dilated shifts of the $\psi^e$.

Recall (see Remark 8.2  in \cite{DHP}) that 
the  tensor product  \eref{mvwave} can
be approximated to accuracy $Cd (2^{-\ell\alpha}+4^{-\ell'})$ by an element from $\Sigma_{C(\ell'+\ell)}$, 
and therefore any function appearing in \eref{wd1} is approximated to this accuracy.   Then, one can obtain comparisons similar to \eref{errorn} between NNA and 
$n$-term wavelet approximation in the multivariate case.  We leave the precise formulation of such comparisons to the reader.

Finally, let us mention briefly a connection of our result with subdivision algorithms
(see \cite{CDM} for a general treatment of these algorithms). We discuss only the univariate  setting. Generally speaking,
a subdivision algorithm begins with a fixed collection $\Lambda=\{(x_i,y_i): i=1,\dots,N\}$  of data points (called {\it control points}), and uses a recursive rule to generate   curves which are then  used for shape design.  In the simplest case, the $n$-th iteration of a subdivision algorithm gives a curve    $y=f_n(x)$ for a suitable function $f_n$.  The rule for generating the $f_n$ can often be described as $f_n=V^ng$, where $g$ is a CPwL function and $V^n$ is the operator associated to a refinement equation. Our results
show that the functions $f_n$ are outputs of neural networks of controlled size.  Therefore, in
these cases, the functions generated by subdivision algorithms lie in these neural network spaces.

\vskip .1in
\noindent
Ingrid Daubechies, Department of Mathematics, Duke University, Durham, NC 27705,\\ ingrid.daubechies@duke.edu

\vskip .1in
\noindent
Ronald DeVore, Department of Mathematics, Texas A\& M University, College Station, TX 77843,\\
rdevore@math.tamu.edu

\vskip .1in
\noindent
Nadav Dym, Department of Mathematics, Duke University, Durham, NC 27705,\\  nadav.dym@duke.edu
\vskip .1in
\noindent
Shira Faigenbaum-Golovin, Department of Mathematics, Duke University, Durham, NC 27705,\\
alexandra.golovin@duke.edu

\vskip .1in
\noindent
 Shahar Z.~Kovalsky, Department of Mathematics, University of North Carolina, Chapel Hill, NC 27599,\\
  shaharko@unc.edu

\vskip .1in
\noindent
Kung-Ching Lin , Department of Mathematics, Texas A\& M University, College Station, TX 77840,\\
kclin@math.tamu.edu

\vskip .1in
\noindent
 Josiah Park, Department of Mathematics, Texas A\& M University, College Station, TX 77840,\\
 j.park@math.tamu.edu

\vskip .1in
\noindent
 Guergana Petrova, Department of Mathematics, Texas A\& M University, College Station, TX 77840,\\
 gpetrova@math.tamu.edu

\vskip .1in
\noindent
Barak Sober, Department of Mathematics, Duke University, Durham, NC 27705,\\
barakino@math.duke.edu

\end{document}